\documentclass[sigconf]{acmart}
\AtBeginDocument{%
  }

\setcopyright{acmlicensed}
\copyrightyear{2018}
\acmYear{xxxx}
\acmDOI{XXXXXXX.XXXXXXX}

\acmConference[ASP-DAC 2025]{xx}{Jan 20--23,2025}{Tokyo, Japan}
\acmISBN{978-1-4503-XXXX-X/18/06}

\usepackage{multirow}
\usepackage{soul}
\usepackage[shortlabels]{enumitem}
\usepackage{array}
\usepackage{hyperref}

\newcolumntype{C}[1]{>{\centering\arraybackslash}p{#1}}

\begin{document}

\title{MICSim: A Modular Simulator for Mixed-signal Compute-in-Memory based AI Accelerator}


\author{Cong Wang, Zeming Chen, Shanshi Huang$^\dag$}
\affiliation{
 \institution{The Hong Kong University of Science and Technology (Guangzhou)
 \city{Guangzhou}
 \country{China}}
 }
 \email{{cwang841, zchen149}@connect.hkust-gz.edu.cn,  shanshihuang@hkust-gz.edu.cn}
\begin{abstract}
This work introduces MICSim, an open-source, pre-circuit simulator designed for early-stage evaluation of chip-level software performance and hardware overhead of mixed-signal compute-in-memory (CIM) accelerators. MICSim features a modular design, allowing easy multi-level co-design and design space exploration. Modularized from the state-of-the-art CIM simulator NeuroSim, MICSim provides a highly configurable simulation framework supporting multiple quantization algorithms, diverse circuit/architecture designs, and different memory devices. This modular approach also allows MICSim to be effectively extended to accommodate new designs.

MICSim natively supports evaluating accelerators’ software and hardware performance for CNNs and Transformers in Python, leveraging the popular PyTorch and HuggingFace Transformers frameworks. These capabilities make MICSim highly adaptive when simulating different networks and user-friendly. This work demonstrates that MICSim can easily be combined with optimization strategies to perform design space exploration and used for chip-level Transformers CIM accelerators evaluation. Also, MICSim can achieve a $9\times\sim 32\times$ speedup of NeuroSim through a statistic-based average mode proposed by this work.
\end{abstract}

\maketitle
\vspace{-0.1cm}

\section{Introduction}
Deep neural networks (DNNs) have been pivotal in the development of artificial intelligence in recent years. Early on, CNN-based neural networks achieved significant breakthroughs in the field of computer vision\cite{lecun2015deep}. More recently, Transformer-based models have made remarkable strides\cite{Vaswani:Transformer,BERT,gpt4}. Nowadays, both CNNs and Transformers are scaling up in size for advanced performance. Besides the powerful model innovations, the explosion of DNNs also owes to the performance improvement of hardware. However, the processors based on the traditional von Neuman architecture, such as GPUs, face throughput and energy efficiency limitations due to extensive data transfers between the process and storage units. This 'memory wall' problem becomes more and more severe as the scale of model parameters increases. A shift toward memory-oriented architectures promises to break the 'memory wall' bottleneck. Compute-in-memory (CIM) technology, which performs in-place multiply-and-accumulation(MAC) within memory crossbars,  minimizing the need for data movement and demonstrating remarkable energy efficiency for deep learning applications \cite{yu2018neuro}. Among different CIM implementations, the mixed-signal CIM implements the MAC operation in the analog domain and usually embraces advanced non-volatile memory(eNVM) devices, characterized by high density and low energy consumption. Impressive energy efficiencies have been reported for mixed-signal CIM acceleration of CNNs \cite{chi2016prime,AEPE,shafiee2016isaac} and Transformers\cite{yiranchen2020retransformer,sridharan2023x,lu2023rram}.

As an emerging architecture, the design of CIM accelerators evolves rapidly and requires cross-layer optimization of algorithms, architecture, circuits, and devices. Therefore, a simulator for early-stage estimation and exploration is necessary before the timing- and resource-consuming deployment. There are already a bunch of simulator platforms\cite{dong2012nvsim,zhu2023mnsim,dnn+neurosim} that have demonstrated powerful capability at system-level or circuit-level CIM accelerator performance evaluation. However, they still have some limitations in terms of generality and flexibility. First, current mainstream CIM pre-circuit simulators (e.g., NeuroSim\cite{dnn+neurosim}, NVSim\cite{dong2012nvsim}) are focused on CNNs. Extending them to the Transformer needs elaborate and time-consuming work due to the tight coupling between the simulator's code structure and the network type. Second, hardware-software co-design techniques like quantization are essential in modern AI accelerator design. While quantization algorithms are evolving rapidly, quickly applying them—whether hardware-specific or not—to CIM hardware is challenging in current simulators. This difficulty arises because these simulators generally treat quantization algorithms as the inherent property, making it hard to incorporate new or updated methods.  Third, current simulators are generally built upon certain architecture/circuit designs, supporting some level of flexibility by parameter tuning. Evaluating new designs or bench-marking across different designs using these simulators needs careful and tedious modification to avoid deviation.

To solve the above issues, we propose MICSim, an open-source, pre-circuit, modular simulator that can provide software and hardware performance estimation for mixed-signal CIM accelerators with different models. The MICSim is available on GitHub (\href{https://github.com/MICSim-official/MICSim_V1.0.git}{https:// github.com/ MICSim-official/MICSim\_V1.0.git}) with main features summarized as follows:
\begin{enumerate}[nosep]
    \item We base MICSim on NeuroSim but unify the construction models of CNNs and Transformers through modular calculations. This allows MICSim to support both without significant modifications. For user-friendliness, we use the popular deep learning framework PyTorch for model training and inference, as well as the widely-used HuggingFace Transformers library for Transformer-based research;
    \item For software performance evaluation, we decouple the modeling of algorithms (quantization), architecture (digit mapping), circuit (ADC), and device (precision, on/off ratio) from each other and provide several typical implementations for easy deployment. Our object-oriented code style enables hardware designers to explore the cross-layer design space and extend to new designs easily;
    \item To further improve the approach-ability of MICSim, we transfer the architecture-level hardware performance into object-oriented Python code. While Python is chosen for easy development, its efficiency is much less than C++. Thus, a data statistic-based based average-mode is employed to replace the trace-based mode to reduce memory usage and runtime. 
  
\end{enumerate}
\vspace{-0.2cm}

\section{MICSim Simulator} 
MICSim integrates the parameter quantization and CIM's hardware effects into the neural network layers, such as Convolution/Linear layers,  from \textit{Pytorch} library\cite{pytorch} in a modular manner. This allows for easy integration of CIM operations into various neural networks to validate the software performance. Specifically, for the Transformers part, to better integrate into the thriving Transformer-based open-source community, MICSim actively embraces the \textit{Transformers} library\cite{HFtransformers} from HuggingFace. The corresponding hardware overhead is formulated with Python-wrapped circuit modules encapsulating the performance models from silicon validated \textit{DNN+NeuroSim V1.3}\cite{NeuroSimv1.3}. 
\vspace{-0.1cm}

\subsection{Software Performance Evaluation}
\label{sec:accuracy}
\begin{figure}
\vspace{-0.4cm}  
\setlength{\abovecaptionskip}{-0.001cm}   
\centering
\includegraphics[width=2.7in]{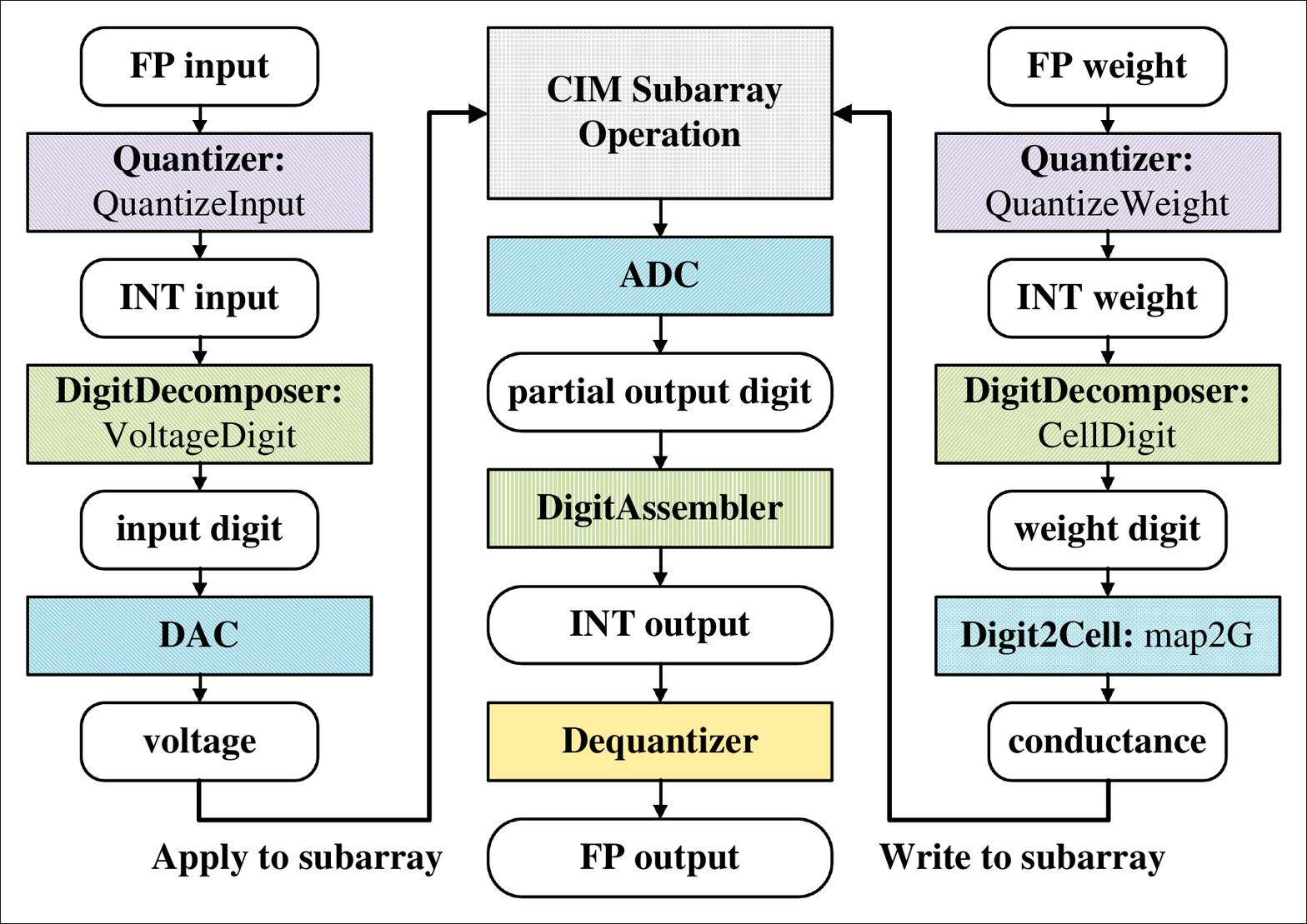}
\caption{The software performance evaluation framework}
\label{fig:accframe}
\vspace{-0.55cm}   
\end{figure}

In the CIM accelerator design, low-precision fixed-point parameters are usually adopted due to the regular structure of the memory array. Also, there will be non-ideal effects, such as limited on/off ratio, cell variation, and ADC quantization loss,  introduced in the calculation by mixed-signal operations. All these hardware-specific effects could degrade the software performance if poorly designed. Thus, we modularized the hardware-induced impact of CIM operation to support the combination of different algorithms, architectures, circuits, and devices. The modular modeling is shown in Fig.\ref{fig:accframe}.

\subsubsection{Quantizer}
\label{sec:quant}
\textbf{Quantizer} is used to quantize weight and input from high-precision floating-point (FP) into low-precision integers (INT) according to the quantization schemes. With an empirical observation, the hardware performance will be better with lower parameter precision. Therefore, quantization is always an important technique in the hardware/software co-design for the CIM accelerator. In MICSim, the input/output data formats of the quantizer are fixed to FP/INT, thus decoupling it from the rest parts of the calculation. A corresponding \textbf{dequantizer} block is introduced at the end of the CIM operation to convert the INT calculation back to the FP representation. Different quantization schemes could be modified to fit into the quantizer/dequantizer module, supporting algorithm exploration despite the circuits and devices implementation. Based on the quantizer/dequantizer definition, MICSim instantiates different quantization algorithms such as dynamic fixed-point (DF)\cite{df}, WAGE\cite{wage} and WAGEUBN\cite{wagev2} etc., as well as the advanced LSQ\cite{lsq} and I-BERT\cite{ibert} algorithms.
\vspace{-0.1cm}

\subsubsection{DigitDecomposer and DigitAssembler}
\label{sec:DigitConverter}
Unlike digital systems that typically adopt the binary number system, operands in CIM subarrays could be high-precision digits ($k$-bit) due to the analog calculation nature. In other words, inputs could be k-bit, represented by $2^k$ voltage levels generated by DACs\cite{shafiee2016isaac}. And weights could be encoded to different conductance values of $1-7$-bit \cite{ankit2019puma} based on the types of memory cells. Although CIM operands can be high-precision, the INT outputs from the quantizer may not be directly mapped to the CIM subarray as their precision ($N$-bit) could be higher than the operands ($k$-bit) supported by the hardware. Thus, we introduce \textbf{DigitDecomposer} to decompose INTs to digits to cover different circuit and device choices. According to the circuit implementation, the conversion from weight INTs to digits could be realized in different ways. In \cite{2compl2019isscc,2compl2020isscc}, weight parameters are decomposed to digits using a 2'complement-like rule. Other works \cite{chi2016prime,cheng2017time} split weight parameters into positive and negative subarrays and thus decompose them into two groups of unsigned values. Unsigned decomposition could also be achieved by shifting weights by a fixed value \cite{shafiee2016isaac}. The output of DigitDecomposer could be directly applied in one CIM subarray operation, decoupling the CIM macro from the digital peripherals on-chip. After the MAC operations, decomposed results will be combined by the \textbf{DigitAssembler} block. While all the methods mentioned above for weights could be applied for input INTs decomposition, binary digits are widely used for inputs to eliminate the hardware cost introduced by DACs. 

\vspace{-0.1cm}

\subsubsection{Digit2Cell and DAC}
Although CIM macros take digits for MAC operation mathematically, the operands are encoded with imperfect analog signals in real implementations. Thus, to mimic the real calculation in the CIM array, we introduce \textbf{Digit2Cell} to leverage imperfection into weight digits. Considering a memory cell with maximum conductance $G_{max}$ and minimum conductance $G_{min}$, to encode $k$-bit weight digits with the memory cell's conductance, the conductance of each cell is divided into $2^k$ levels between $G_{max}$ and $G_{min}$. Thus, a weight digit ${d_w^i}$ will be mapped to cell conductance $G^i$ as below:
\begin{equation}
G^i = {d_w^i} \times \Delta G + G_{\text{min}}, \quad {d_w^i} \in [0, 2^k - 1], \quad \Delta G = \frac{G_{\text{max}} - G_{\text{min}}}{2^k - 1}
\end{equation}
The conductance is not linearly proportional to ${d_w^n}$ due to the $G_{min}$ term, which introduces non-idealities into the CIM operation. In the Digit2Cell module, we normalize the conductance, generalizing weight digit modeling across different devices while maintaining the non-idealities. Thus, weight digit ${d_w^i}$ involved in CIM crossbar's analog computation is changed to ${d_{cell}^i}$ as equation (\ref{eq:wcell}). In addition, our modeling also takes into account the cell variation by adding random noise to the normalized weight digit.

\begin{align}
\label{eq:wcell}
 d_{cell}^i =\frac{G^i}{\Delta G} = {d_w^i} + \frac{G_{min}}{\Delta G} 
\end{align}

Similarly, we introduce a DAC block to convert the input digits to read voltage and normalize it back to integrate possible non-idealities. The normalization of Digit2Cell and DAC is inherent to the CIM operations by correspondingly picking references for the ADC. 

\subsubsection{ADC}
\textbf{ADC} converts the analog MAC results into digital signals. In this digitization process, the quantization loss introduced should be considered. A general process of ADC can be mathematically formulated as a piece-wise function that maps its input to corresponding centers based on configured edges. For a $k$-bit ADC, it can be represented as:
\begin{align}
\label{eq:adc}
\text{f}(x) = 
\begin{cases} 
c_0 & \text{if } x < \text{ref}_0 \\
c_i & \text{if } \text{ref}_{i-1} \leq x < \text{ref}_i, \text{ for } i = 1, \ldots, 2^k - 2 \\
c_{2^k - 1} & \text{if } x \geq \text{ref}_{2^k - 2}
\end{cases}
\end{align}
where $c_i$ represents the quantized value of each level, and $\text{ref}_i$ refers to the reference that defines the edge of each level in the ADC.

This general format could represent any ADC by projecting $c_i$ and $\text{ref}_i$. MICSim offers two general ADC choices: linear and nonlinear. Linear ADCs have fixed step size between $c_{i-1}$ and $c_i$, while nonlinear ADCs determine their $c_i$ and $\text{ref}_i$ based on the statistical distribution of the data. Our modeling of ADC is based on (\ref{eq:adc}) with user-defined $c_i$ and $\text{ref}_i$. Thus, it offers great flexibility for users to test their own ADCs without being confined to the existing implementations.
\vspace{-0.2cm}

\subsection{Hardware Performance Evaluation}
MICSim offers hardware performance evaluation based on a chip architecture similar to NeuroSim. Unlike NeuroSim, which focuses on evaluations of different devices and technology nodes, MICSim is designed for easy implementation of a CIM accelerator with the combination of different algorithms, architecture, circuits, and devices. Thus, flexibility is a key consideration in MICSim's design. By wrapping the circuit modules from NeuroSim, MICSim enables chip-level hardware performance evaluation in an object-oriented and hierarchical manner using Python.

\begin{figure}
\vspace{-0.4cm}  
\setlength{\abovecaptionskip}{-0.05cm}   
    \centering
    \includegraphics[width=3in]{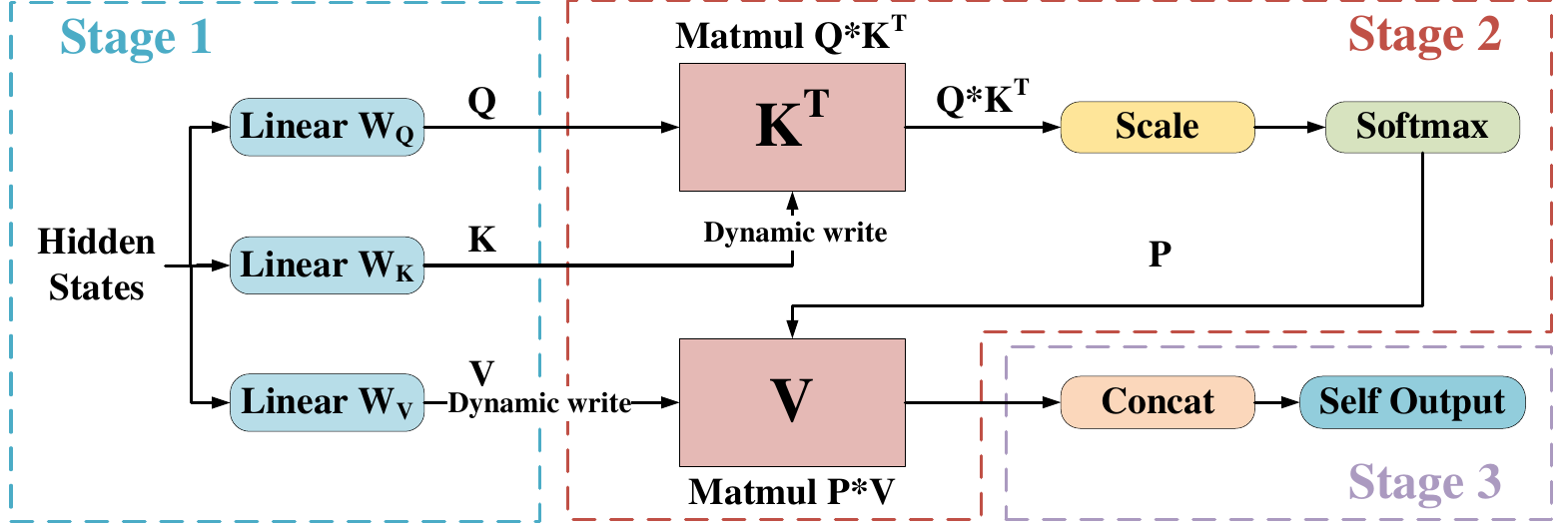}
    \caption{Dataflow in multi-head attention block}
    \label{fig:3stages}
\vspace{-0.55cm}       
\end{figure}

\subsubsection{Chip Architecture}
\label{sec:chiparch}
\begin{figure}
\vspace{-0.45cm}  
\setlength{\abovecaptionskip}{-0.1cm}   
\centering
\includegraphics[width=2.6in]{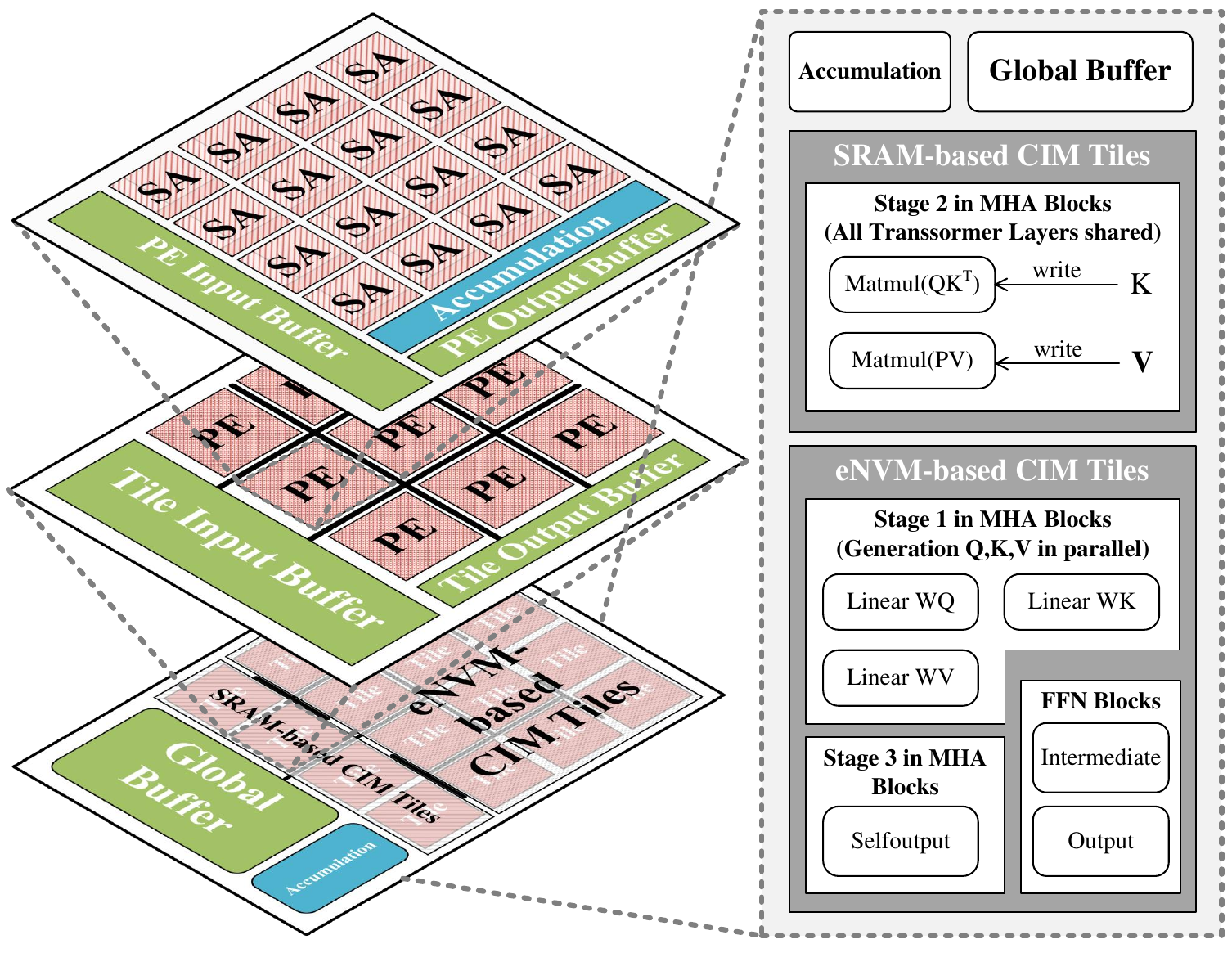}
\caption{Transformer CIM accelerator chip architecture}
\label{fig:chip}
\vspace{-0.6cm} 
\end{figure}

Inheriting the hierarchical chip design from NeuroSim, MICSim features a three-layer architecture consisting of Subarray (SA), Processing Engine (PE), and Tile, shown in the left-hand side of Fig.\ref{fig:chip}. This architecture is designed to maximize data reuse for CNNs, but it may not always be efficient for different networks or structures. To address this, MICSim unified the definitions of SA, PE, and Tile, treating all digital circuit modules as objects. This approach allows a chip to be assembled like LEGO, facilitating easy architecture optimization of different networks or design space explorations for various tasks.

MICSim originally supports the hardware evaluation for CNN and Transformer. The default CNN accelerator design is similar to NeuroSim. For CIM accelerators targeting Transformers, they focus on accelerating multi-head attention(MHA) blocks(Fig\ref{fig:3stages}) and the subsequent feed-forward(FFL) blocks. These two consecutive blocks involve intensive matrix multiplication operations, which can be categorized into static matrix multiplication(SMM) and dynamic matrix multiplication (DMM) based on whether they require dynamic writing into the memory subarrays.

Fig.\ref{fig:chip} illustrates the architecture of Transformer CIM accelerators adopted by MICSim, which employs heterogeneous tiles on-chip. For SMM accelerations, eNVM-based CIM tiles are adopted since the weights are well-trained neural network parameters. For DMM in Transformers, SRAM-based CIM tiles are preferred because the weights are generated on-the-fly for each input. The right-hand side of Fig.\ref{fig:chip} shows the mapping strategy applied. The two-layer FFL blocks, Stage1 and Stage 3 in MHA blocks (Fig\ref{fig:3stages}) are SMM, mapped to eNVM-based CIM tiles. The Stage2 in MHA block includes two DMM: $Matmul(\mathbf{Q}, \mathbf{K}^T)$ and $Matmul(\mathbf{P}, \mathbf{V})$. For $Matmul(\mathbf{Q}, \mathbf{K}^T)$, each column of the generated $\mathbf{K}$ matrix is written into each column of the SRAM-based CIM array as weights. Due to the characteristics of CIM computation, the transposition operation is naturally implemented by the weight loading. Then, for the computation of $Matmul(\mathbf{P}, \mathbf{V})$, it is feasible to write the $\mathbf{V}$ matrix into the SRAM-based CIM array while computing matrix $\mathbf{P}$, thereby achieving higher parallelism. The latency generated by writing $\mathbf{V}$ can be hidden in this case.

\vspace{-0.1cm}
\subsubsection{Average Data Mode}
\label{sec:average}

According to \cite{dnn+neurosim}, the data pattern of the weight/input involved in the computation highly affects the energy consumption and latency of CIM operation. A crucial feature of NeuroSim for accurate hardware overhead estimation is the trace mode, which calculates the latency and energy consumption based on the real data involved in the computation. NeuroSim dumps out real-traced weight/input values during the inference and iterates through them for hardware overhead evaluation. However, this method demands a long runtime and substantial memory usage during simulation, particularly for large models of Transformers. To address this challenge, we propose the average mode, which leverages the statistics of the weight and input data instead of the entire trace.

\begin{figure}
\vspace{-0.3cm}  
\setlength{\abovecaptionskip}{-0.05cm}   
    \centering
    \includegraphics[width=2.6in]{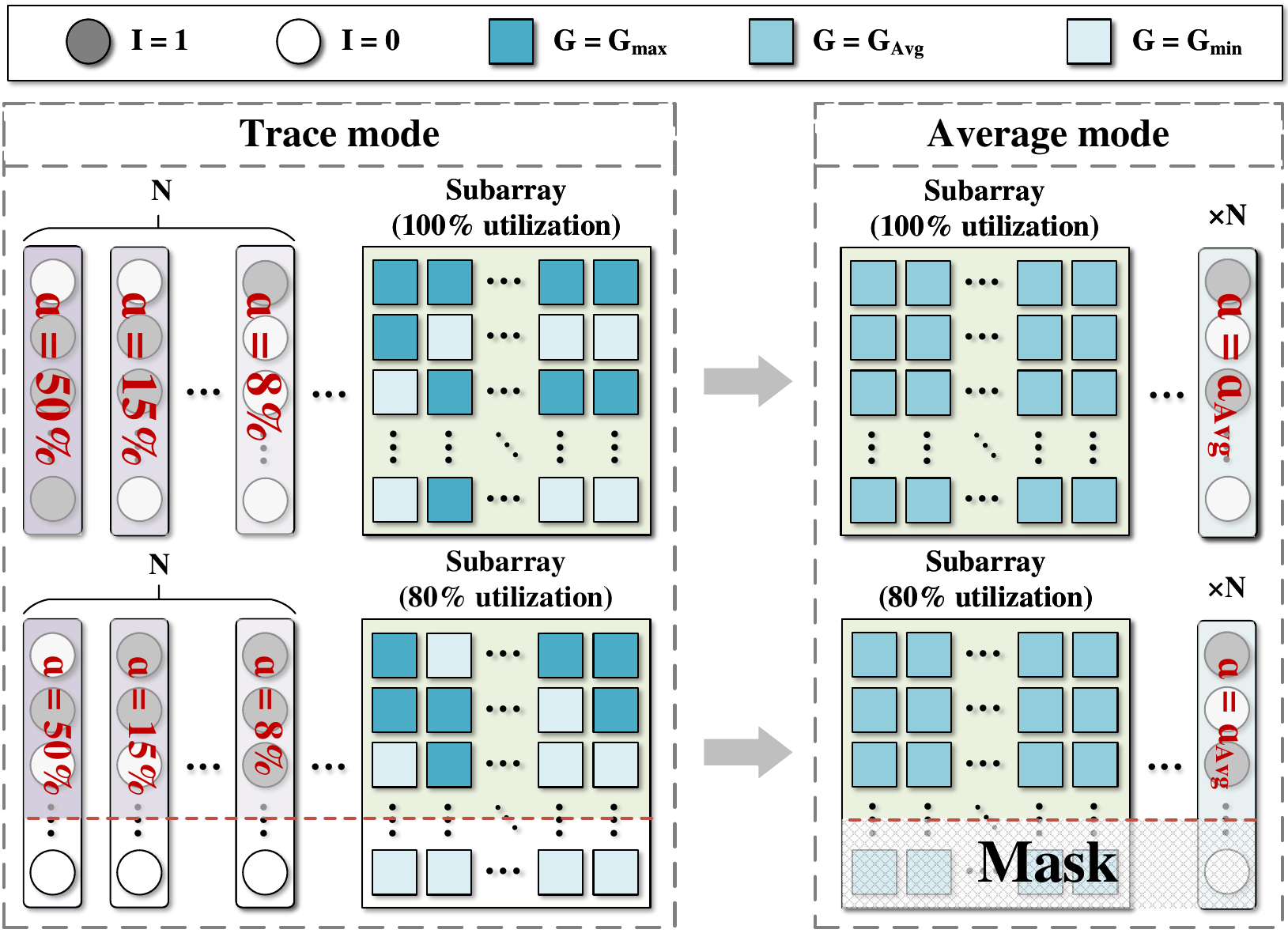}
    \caption{Comparison of average and trace mode workflows}
    \label{fig:averagemode_workflow}
\vspace{-0.65cm}       
\end{figure}

Fig.\ref{fig:averagemode_workflow} conceptually illustrates the average mode and its difference from the trace mode. For binarized inputs applied to CIM arrays, $\alpha$ is introduced to represent the ratio of non-zero elements in an input vector. Thus, the data pattern of inputs is specifically reflected by the $\alpha$ of each input vector. Also, illustrated by a 1-bit cell case, the data pattern of the weights is encoded by different cell conductance. For a certain subarray, the trace mode needs to process all of its $n$ input vectors and accumulate the results to obtain the total latency and energy introduced by this subarray. As weight traces differ for each subarray, this $n$-times iteration must be applied to each subarray.

Statistics of the data, instead of the trace, are obtained during inference in the average mode. Specifically, a set of average data is generated for each layer in network to represent the input/weight data pattern of that layer: $\alpha_{avg}$ is the average ratio of non-zero elements in inputs, and $G_{avg}$ is used to represent the average conductance of weights in all subarrays of that layer. In principle, for a certain layer, the total hardware overhead could be calculated by 1) applying one input vector with non-zero elements of $\alpha_{avg}$ to one subarray of $G_{avg}$; 2) scaling the result of step 1) by the number of inputs $n$ and the number of subarrays. However, based on the weight shape and subarray size, the utilization of all subarrays in a layer is not completely $100\%$. For each subarray, the real statistics should be corrected by a mask generated from memory utilization. Thus, we still need to iterate each subarray in a layer while applying only one masked input for each subarray. As a result, the average mode MICSim uses is sped up by $n$ times compared to the trace mode. To further speed up the calculation, we combine subarrays with same memory utilization and calculate them once.
\vspace{-0.2cm}

\section{Experiment Results}
In this section, we first validate the performance of the average mode and demonstrate its acceleration capabilities. Then, we conduct a design space exploration using MICSim to showcase the benefits of our modular approach. Finally, we present the evaluation of accelerator for transformer-based models using MICSim.
\vspace{-0.5cm}

\subsection{Average mode Versus Trace mode}
\subsubsection{Performance}

As discussed in Section \ref{sec:average}, we adopt the average mode based on data statistics to mimic the trace mode of NeuroSim while improving the calculation efficiency. It is worth noting that the data pattern will not impact the chip area evaluation but will affect the subarray's read latency and dynamic energy. To validate the performance of the average mode, we evaluate hardware overhead with the same chip configuration for the two modes. Table \ref{tab:avg_networks} compares throughput and energy efficiency for different networks between the average mode and the trace mode. It can be seen that our average mode has a minimal impact on throughput compared to the trace mode, with less than $10\%$ differences in energy efficiency. 

\begin{table}[h]
\centering
\small
\begin{tabular}{|c|c|c|c|c|}
\hline
\textbf{Index} & \textbf{Mode} & \textbf{VGG8} & \textbf{ResNet18} & \textbf{DenseNet40} \\
\hline
\multirow{2}{*}{\shortstack{Energy Efficiency \\ (TOPS/W)}} & Trace & 10.01 & 5.62 & 6.22 \\
 & Average & 9.27 & 5.19 & 5.67 \\
\hline
\multirow{2}{*}{\shortstack{Throughput (TOPS)}} & Trace & 1.54 & 0.46 & 0.11 \\
 & Average & 1.54 & 0.46 & 0.11 \\
\hline
\end{tabular}
\caption{Comparsion of average and trace mode among various netwrok}
\label{tab:avg_networks}
\vspace{-0.85cm} 
\end{table}


\vspace{-0.1cm}
\subsubsection{Runtime}
\begin{figure}
\vspace{-0.2cm}  
\setlength{\abovecaptionskip}{-0.1cm}   
    \centering
    \includegraphics[width=2.7in]{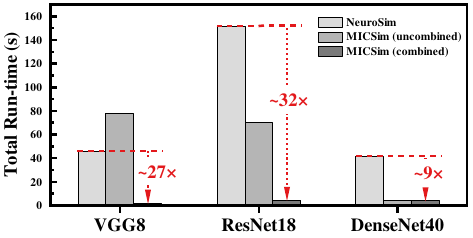}
    \caption{Runtime comparison between NeuroSim and MICSim}
    \label{fig:runtime}
\vspace{-0.65cm}    
\end{figure}

Fig.\ref{fig:runtime} shows the runtime comparison between MICSim and NeuroSim for hardware performance evaluation. The average mode in MICSim has less computational load than the trace mode in principal. However, since the MICSim is built in Python for chip-level performance evaluation, the loop through different subarrays is much slower than the C++ loop in NeuroSim. As a result, when evaluating networks with large filters (thus numerous subarrays for weights) but small input feature maps, such as VGG8, MICSim performs even slower than NeuroSim. With the speedup technique of combining subarrays with same memomry untilization, as described in Section \ref{sec:average}, MICSim improves the estimation of the CNNs up to $9\sim 32$ times faster than NeuroSim despite the slower programming language, as shown in Fig.\ref{fig:runtime}. This significant enhancement demonstrates MICSim's efficiency and capability in accelerating mixed-signal compute-in-memory simulations.

\vspace{-0.2cm}

\subsection{Design space exploration using MICSim}
MICSim enables cross-layer design space exploration of algorithms (quantization), architectures (digit mapping), circuits (ADC), and devices (cell precision, on/off ratios). The decoupled and modularized modeling of these elements allows easy configuration of different designs or extensions to new designs. This section conceptually demonstrates the design space exploration using MICSim for a CNN inference accelerator. For simplicity, we adopt a greedy algorithm for space searching. 

\subsubsection{Algorithm level}
MICSim supports both post-training quantization(PTQ) and quantization-aware training(QAT) of different algorithms in-situ. In this stage, INT output will directly be used for matrix multiplication without introducing any architecture/circuit/ device effect. Thus, the only impact factor on the software performance will be quantization. This work explores three QAT algorithms for CNN: WAGE, DF, and LSQ, on two CNN networks, VGG8 for Cifar10\cite{cifar10} and ResNet18 for ImageNet(Subset)\cite{imagenet}. Table \ref{tab:quantization} shows the minimum precision needed for each network under different quantization algorithms. The quantization precision for inputs and weights is consistent, considering the simplicity of demonstration, but it could be different in practice to improve the performance further.  

\begin{table}[h]
\centering
\small
\begin{tabular}{|c|c|c|c|}
\hline
\textbf{Quantization} & \textbf{WAGE} & \textbf{LSQ} & \textbf{DF} \\
\hline
\multicolumn{4}{|c|}{\textbf{Minimum parameter}} \\
\hline
VGG8 (CIFAR10) & 3-bit & 3-bit & 4-bit \\
\hline
ResNet18 (ImageNet) & 7-bit & 4-bit & 5-bit  \\
\hline
\multicolumn{4}{|c|}{\textbf{Hardware perforamcne}} \\
\hline
\multicolumn{4}{|c|}{VGG8 on Cifar10} \\
\hline
Area (mm\(^2\)) & {33.56} & {33.56
}& {41.01} \\
\hline
Energy Efficiency (TOPS/W) & {36.57} & {33.34} & {32.93} \\ 
\hline
Throughput (TOPS) & {2.54} & {2.54} & {2.27} \\
\hline
\multicolumn{4}{|c|}{ResNet18 on Imagenet} \\ 
\hline
Area (mm\(^2\)) &{147.04} & {96.13} & {115.24} \\
\hline
Energy Efficiency (TOPS/W) & {4.08} & {18.57} & {12.39} \\ 
\hline
Throughput (TOPS) & {0.24} & {0.37} & {0.29} \\
\hline
\end{tabular}
\caption{Minimum parameters and performance comparison among different quantization method}
\label{tab:quantization}
\vspace{-1cm} 
\end{table}


Considering a unified hardware configuration but parameter precision that satisfies all the quantization optimization, the hardware performance of corresponding accelerators is shown in Table \ref{tab:quantization}. In conclusion, WAGE with $3$-bit precision achieves the best hardware performance for VGG8. Among the three quantization methods for VGG8, $4$-bit DF shows the worse hardware performance due to a higher parameter precision. WAGE and LSQ have the same area and similar throughput since they can both achieve $3$-bit. However, the energy efficiency of LSQ is lower than that of WAGE, primarily because different quantization algorithms differ in data statistics of weights and inputs, thereby affecting energy consumption. For ResNet18, LSQ, requiring the lowest input and weight precision, performs best in all metrics. Therefore, in the following process, we fix the quantization method to $3$-bit WAGE for VGG8 and $4$-bit LSQ for ResNet18.


\subsubsection{Architecture and Circuit level}
\label{sec:adc}
\begin{figure}
\setlength{\abovecaptionskip}{-0.05cm}   
    \centering
    \includegraphics[width=3.4in]{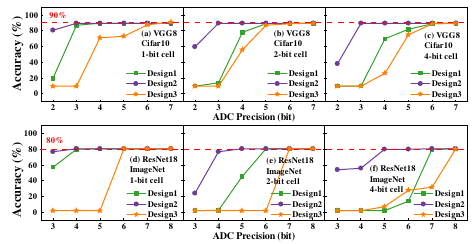}
    \caption{Software performance with various ADC precision under different circuit architecture design}
    \label{fig:adc_acc}
    \vspace{-0.5cm} 
\end{figure}

As discussed in Section \ref{sec:DigitConverter}, MICSim supports designs with different architectures for digit mapping, which not only affect software performance but also impact the hardware performance of the chip. For simplicity, we refer to the three types of circuit architecture from Section \ref{sec:DigitConverter} as Design1 (2’complement-like), Design2 (positive-negative split), and Design3 (weight shifted). ADC and memory cell precision are also considered while exploring architecture as they are highly coupled. However, an infinite on/off ratio is assumed to eliminate the impact of a certain memory cell. Generally, we prefer lower ADC precision for better hardware performance\cite{shimeng2021review}. Fig.\ref{fig:adc_acc} shows the software performance across different settings. For both VGG8 and ResNet18, Design2 always requires the lowest ADC precision compared to the other two options, while Design3 performs the worst, needing higher ADC precision in all cases. Design1's ADC precision requirement is similar to Design2's when cell precision is low, but it increases as cell precision increases. 


\begin{figure}
\setlength{\abovecaptionskip}{-0.05cm}   
    \centering
    \includegraphics[width=3.4in]{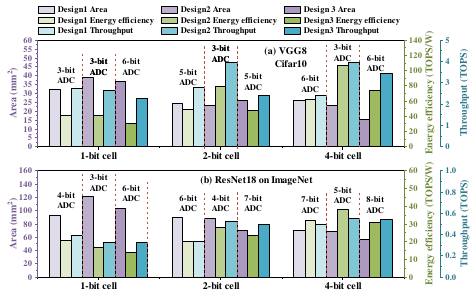}
    \caption{Hardware performance among different circuit architecture design }
    \label{fig:adc_per}
    \vspace{-0.7cm} 
\end{figure}



To evaluate the hardware performance, we finalize the ADC precision settings under a $3\%$ accuracy degradation tolerance for different cases (Fig.\ref{fig:adc_acc}). RRAM cells with Ron/Roff = $6k\Omega / 900k \Omega$ \cite{rram150} are assumed for hardware performance evaluation, and the results are shown in Fig.\ref{fig:adc_per}. We can see that when cell precision is $1$, Design2 has the largest area overhead. This is because Design2 uses a pair of subarrays to store signed weights, while the other two only need a single array. However, despite having subarray pairs, the area of Design2 is not a double of Design1 or Design3, thanks to the area saved by low precision ADCs. As cell precision increases to $4$ bits, the area of Design2 even becomes smaller than Design1. Aside from the difference in ADC precision, Design1 falls behind in high cell precision cases because it needs to store the sign bit separately.  It is interesting to see that, despite having higher ADC precision, Design3 achieves the smallest area when cell precision is $4$. Compared to the subarray pair needed by Design2 or the extra sign bit introduced by Design1, Design3 represents sign weights at the cost of dummy columns, which introduces a much smaller area overhead. Due to Design2's property to tolerate more aggressive ADC precision reduction at higher cell precision, Design2 achieves significantly higher energy efficiency and throughput when the cell bit is $4$. Since energy efficiency is usually the priority in the CIM design, we chose Design2 at the architecture level with a circuit design of $3$-bit ADC for VGG8 and $5$-bit for ResNet18, despite the area cost.



\begin{table*}
\centering
\small
\begin{tabular}{|c| c|c|c | c|c|c | c|c|c | c|c|c | c|c|c|}
\hline
\textbf{Device} & \multicolumn{3}{c|}{\textbf{RRAM\cite{rram150}}} & \multicolumn{3}{c|}{\textbf{RRAM\cite{jain201913}}} & \multicolumn{3}{c|}{\textbf{RRAM\cite{wu2018methodology}}} & \multicolumn{3}{c|}{\textbf{FeFET\cite{ni2018fefet}}} & \multicolumn{3}{c|}{\textbf{PCM\cite{kim2019PCM}}} \\
\hline
Ron$(\Omega)$ & \multicolumn{3}{c|}{6k} & \multicolumn{3}{c|}{6k} & \multicolumn{3}{c|}{100k} & \multicolumn{3}{c|}{240k} & \multicolumn{3}{c|}{40k} \\

\hline

On/Off & \multicolumn{3}{c|}{150} & \multicolumn{3}{c|}{17} & \multicolumn{3}{c|}{10} & \multicolumn{3}{c|}{100} & \multicolumn{3}{c|}{12.5}\\
\hline
Cell precision  & 1 & 2 & 4 & 1 & 2 & 4 & 1 & 2 & 4 & 1 & 2 & 4 & 1 & 2 & 4 \\
\hline
{VGG8, baseline:90\%}
& 90\% & 89\% & 89\% 
& 88\% & 86\% & 10\% 
& 88\% & 75\% & 10\%
& 90\% & 89\% & 88\% 
& 89\% & 82\% & 10\% \\
\hline
{ResNet18, baseline:80\%} 
& 81\% & 81\% & 78\% 
& 81\% & 81\% & 14\% 
& 81\% & 81\% & 3\% 
& 81\% & 81\% & 78\% 
& 81\% & 81\% & 2\% \\
\hline
\end{tabular}
\caption{Software performance among different devices}
\label{tab:device—acc}
\vspace{-0.7cm} 
\end{table*}

\subsubsection{Device level}
\label{sec:final}
\begin{figure}
\vspace{-0.1cm}  
\setlength{\abovecaptionskip}{-0.01cm}   
    \centering
    \includegraphics[width=2.7in]{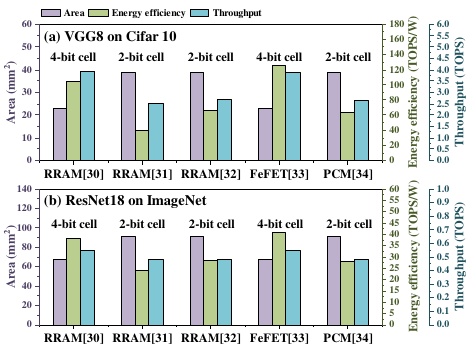}
    \caption{Hardware performance under memory cells}
    \label{fig:deviceperformance}
    \vspace{-0.65cm} 
\end{figure}

With the accelerator configured from the previous step, we explore some device candidates listed in Table \ref{tab:device—acc}. To recapitulate, VGG8 adopts $3$-bit input/weights quantized by WAGE while ResNet18 adopts 4-bit input/weights quantized by LSQ. Both employ the architecture based on Design2, with ADC of $3$-bit and $5$-bit, respectively. Table \ref{tab:device—acc} shows that with a $3\%$ accuracy drop tolerance, only RRAM\cite{rram150} and FeFET\cite{ni2018fefet} can maintain accuracy at $4$-bit precision. The remaining memory cells can only maintain accuracy at a lower bit precision due to the limited on/off ratio. The hardware performance under each memory cell with a precision satisfying accuracy requirement is shown in Fig.\ref{fig:deviceperformance}. We can see that the FeFET\cite{ni2018fefet} with 4-bit precision demonstrates the highest energy efficiency for these two networks.
\vspace{-0.1cm}

\subsubsection{Discussion}
The above experiments demonstrate a conceptual process for design space exploration using MICSim. It could be seen that while MICSim decouples the design space into different levels for easy configuration and extension, each sub-level is highly coupled with the others in determining the final software/hardware performance. Thus, we do not explore the design space strictly sequentially for each sub-level. However, the hardware design choices obtained in Section \ref{sec:final} could still be sub-optimal. Due to the simplicity of configuring MICSim and the short runtime, MICSim could also be integrated into search algorithms such as genetic algorithms or simulated annealing.

\vspace{-0.2cm}
\subsection{Evaluation of Transformers}

In this section, we demonstrate the evaluation of software and hardware performance of the CIM-based accelerators for Transformers using MICSim. The BERT (base) model\cite{BERT} is fine-tuned on the SST-2 task from the GLUE dataset\cite{glue}. All weights and inputs of the model are quantized to $8$-bit by the QAT method from the I-BERT\cite{ibert} method, achieving a baseline accuracy of $90.02\%$. An RRAM with Ron/Roff = $6k\Omega / 900k \Omega$ with 2-bit per cell\cite{rram150} is assumed for hardware performance evaluation. The other evaluation settings and results are summarized in Table \ref{tab:transper}. We observe that Design2 needs the least ADC precision, while the ADC requirement of Design3 is strict. This trend is consistent with the conclusions in Section \ref{sec:adc}, even though the accelerated model has shifted from CNNs to Transformers. Under these settings, we can see that Design2 has the largest area but achieved the best energy efficiency. In terms of throughput, Design3 achieved the highest throughput, while Design2 is the worst. 

\begin{table}[ht]
    \centering
    \small
    \begin{tabular}{|c|c|c|c|}
        \hline
        \textbf{Technology node} & \multicolumn{3}{c|}{22nm} \\
        \hline
        \multirow{2}*{\textbf{Subarray Size}} & \multicolumn{3}{c|}{64 x 64 SRAM subarray (DMM)} \\
        & \multicolumn{3}{c|}{128 x 128 RRAM subarray (SMM)} \\
        \hline
        \textbf{Circuit architecture} & Design 1 & Design 2 & Design 3 \\
        \hline
        \textbf{SAR-ADC precision} & 8 & 7 & 9 \\
        \hline
        \multirow{1}*{\textbf{Accuracy}} & \multicolumn{3}{c|}{$90.02\%$} \\
        \hline
        \textbf{Area (mm\(^2\))} & {147.565} & {204.878} & {122.452} \\
        \hline
        \textbf{Energy efficiency (TOPS/W)} & {4.163} & {4.554} & {3.658} \\
        \hline
        \textbf{Throughput (TOPS)} & {0.126} & {0.116} & {0.135} \\
        \hline
        \textbf{Compute efficiency (TOPS/mm\(^2\))} & {0.000855} & {0.000564} & {0.001102} \\
        \hline
    \end{tabular}

    \caption{Benchmark results of CIM accelerator chips for Bert among different circuit architecture}
    \label{tab:transper}
    \vspace{-0.8cm} 
\end{table}

\begin{figure}
\centering
\vspace{-0.2cm}  
\setlength{\abovecaptionskip}{-0.05cm}   
\includegraphics[width=3.4in]{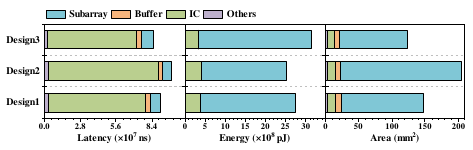}
\caption{Breakdown of overhead}
\label{fig:breakdown}
\vspace{-0.6cm} 
\end{figure}

Fig.\ref{fig:breakdown} provides a breakdown of the hardware overhead of these three designs. We observe that the proportions of overhead for various parts (subarray, buffer, interconnect network (IC)) on the chip are similar across different design choices. Specifically, subarrays contribute to the majority of the area and energy overhead, while internal data movement on chip results in the IC being the primary source of latency overhead. The underlying reason for differences in their area and energy footprint is consistent with the analysis in Section \ref{sec:adc}. Latency difference is mainly caused by area difference as a larger area requires a longer distance for data transfer, and thus a longer IC delay. Hence, Design3 with the smallest area can reach the highest throughput, while the largest Design2 gets the lowest throughput.

\vspace{-0.2cm}
\section{Conclusion}
In this work, we present MICSim, a simulator designed to evaluate the software performance and hardware overhead of CIM accelerators for CNNs and Transformers. MICSim introduces the average mode, which reduces runtime and memory usage while maintaining accurate hardware overhead estimations. With its modular design, MICSim supports multiple quantization algorithms, and various circuit
architecture designs. It could be further extended to new implementations and designs easily or be integrated into optimization strategies to effectively performs design space exploration.
\vspace{-0.2cm}

\bibliographystyle{ACM-Reference-Format}



\end{document}